\documentclass[]{llncs}
\pdfoutput=1
\usepackage{amsmath}
\usepackage{amsfonts}
\usepackage{verbatim}
\usepackage[usenames]{color}
\usepackage{graphicx}
\usepackage{stmaryrd}
\usepackage[mathscr]{eucal}
\usepackage{ifthen}
\input xy
\xyoption{all}
\newboolean{showProofs}
\newtheorem{clam}{Claim}
\newcommand{\p}[0]{p^{\phantom{g}}}

\newcommand{\ranglel}[0]{\rangle^{\phantom{j}}}

\setboolean{showProofs}{true}
\title{Sufficient Conditions for Coarse-Graining  Evolutionary Dynamics
}
\author{Keki Burjorjee}
 \institute{DEMO Lab,\\ Computer Science Department,\\ Brandeis University, Waltham, MA 02454\\kekib@cs.brandeis.edu}
\begin{document}

\date{}
\maketitle
\begin{abstract}
It is commonly assumed that the ability to track the frequencies of a set of schemata in the evolving population of an infinite population genetic algorithm (IPGA) under different fitness functions will advance efforts to obtain a theory of adaptation for the simple GA. Unfortunately, for IPGAs with long genomes and non-trivial fitness functions there do not currently exist theoretical results that allow such a study.  We develop a simple framework for analyzing the dynamics of an infinite population evolutionary algorithm (IPEA). This framework derives its simplicity from its abstract nature. In particular we make no commitment to the data-structure of the genomes, the kind of variation performed, or the number of parents involved in a variation operation. We use this framework to derive abstract conditions under which the dynamics of an IPEA can be \emph{coarse-grained}. We then use this result to derive concrete conditions under which it becomes computationally feasible to closely approximate the frequencies of a family of schemata of relatively low order over multiple generations, even when the bitstsrings in the evolving population of the IPGA are long.
\end{abstract}

\section{Introduction}
It is commonly assumed that theoretical results which allow one to track the frequencies of schemata in an evolving population of an infinite population genetic algorithm (IPGA) under different fitness functions will lead to a better understanding of how GAs perform adaptation \cite{holland75:_adapt_natur_artif_system,Goldberg:1989:GAS,Mitchell:1996:IGA}. An IPGA with genomes of length $\ell$ can be modelled by a set of $2^\ell$ coupled difference equations. For each genome in the search space there is a corresponding state variable which gives the frequency of the genome in the population, and a corresponding difference equation which describes how the value of that state variable in some generation can be calculated from the values of the state variables in the previous generation. A naive way to calculate the frequency of some schema over multiple generations is to numerically iterate the IPGA over many generations, and for each generation, to sum the frequencies of all the genomes that belong to the schema.  The simulation of one generation of an IPGA with a genome set of size $N$ has time complexity $O(N^3)$, and an IPGA with bitstring genomes of length $\ell$ has a genome set of size $N=2^\ell$. Hence, the time complexity for a  numeric simulation of one generation of an IPGA is $O(8^\ell)$ . (See \cite[p.36]{vose:1999:sgaft} for a description of how the Fast Walsh Transform  can be used to bring this bound down to $O(3^\ell)$.) Even when the Fast Walsh Transform is used, computation time still increases exponentially with $\ell$. Therefore for large $\ell$ the naive way of calculating the frequencies of schemata over multiple generations clearly becomes computationally intractable\footnote{Vose reported in 1999 that computational concerns force numeric simulation to be limited to cases where $\ell\leq20$}.

Holland's schema theorem \cite{holland75:_adapt_natur_artif_system,Goldberg:1989:GAS,Mitchell:1996:IGA} was the first theoretical result which allowed one to calculate (albeit imprecisely) the frequencies of schemata after a single generation. The crossover and mutation operators of a GA can be thought to destroy some schemata and construct others. Holland only considered the destructive effects of these operators. His theorem was therefore an inequality. Later work \cite{conf/icga/StephensW97} contained a theoretical result which gives exact values for the schema frequencies after a single generation. Unfortunately for IPGAs with long bitstrings this result does not straightforwardly suggest conditions under which schema frequencies can be numerically calculated over multiple generations in a computationally tractable way.

\subsection{The Promise of Coarse-Graining}
Coarse-graining is a technique that has widely been used to study aggregate properties (e.g. temperature) of many-body systems with very large numbers of state variables (e.g. gases). This technique allows one to reduce some system of difference or differential equations with many state variables (called the fine-grained system) to a  new system of difference or differential equations that describes the time-evolution of a smaller set of state variables (the coarse-grained system). The state variables of the fine-grained system are called the microscopic variables and those of the coarse-grained system are called the macroscopic variables. The reduction is done using a surjective non-injective function between the microscopic state space and the macroscopic state space called the partition function. States in the microscopic state space that share some key property (e.g. energy) are projected to a single state in the macroscopic state space. The reduction is therefore `lossy', i.e. information about the original system is typically lost. Metaphorically speaking, just as a stationary light bulb projects the \emph{shadow} of some moving 3D object onto a flat 2D wall, the partition function projects the changing state of the fine-grained system onto states in the state space of the coarse-grained system.

The term `coarse-graining' has been used in the Evolutionary Computation literature to describe different sorts of reductions of the equations of an IPGA. Therefore we now clarify the sense in which we use this term. In this paper a reduction of a system of equations must satisfy three conditions to be called a coarse-graining. Firstly, the number of macroscopic variables should be smaller than the number of microscopic variables. Secondly, the new system of equations must be completely self-contained in the sense that the state-variables in the new system of equations must \emph{not} be dependent on the microscopic variables. Thirdly,  the dynamics of the new system of equations must `shadow' the dynamics described by the original system of equations in the sense that if the projected state of the original system at time $t=0$ is equal to the state of the new system at time $t=0$ then at any other time $t$, the projected state of the original system should be closely approximated by the state of the new system. If the approximation is instead an equality then the reduction is said to be an \emph{exact} coarse-graining. Most coarse-grainings are not exact. This specification of  coarse-graining is consistent with the way this term is typically used in the scientific literature. It is also similar to the definition of coarse-graining given in \cite{journals/tcs/RoweVW06} (the one difference being that in our specification a coarse-graining is assumed not to be exact unless otherwise stated).

Suppose the vector of state variables $\mathbf x^{(t)}$ is the state of some system at time $t$ and the vector of state variables $\mathbf y^{(t)}$ is the state of a coarse-grained system at time $t$.  Now, if the partition function projects $\mathbf x^{(0)}$ to $\mathbf y^{(0)}$, then, since none of the state variables of the original system are needed to express the dynamics of the coarse-grained system, one can determine how the state of the coarse-grained system $\mathbf y^{(t)}$ (the shadow state) changes over time without needing to determine how the state in the fine-grained system $\mathbf x^{(t)}$ (the shadowed state) changes. Thus, even though for any $t$, one might not be able to determine $\mathbf x^{(t)}$, one can always be confident that $\mathbf y^{(t)}$ is its projection. Therefore, if the number of state variables of the coarse-grained system is small enough, one can numerically iterate the dynamics of the (shadow) state vector $\mathbf y^{(t)}$ without needing to determine the dynamics of the (shadowed) state vector $\mathbf x^{(t)}$.

In this paper we give sufficient conditions under which it is possible to coarse-grain the dynamics  of an IPGA such that the macroscopic variables are the frequencies of the  family of schemata in some schema partition. If the size of this family is small then, regardless of the length of the genome, one can use the coarse-graining result to numerically calculate the approximate frequencies of these schemata over multiple generations in a computationally tractable way. Given some population of bitstring genomes, the set of frequencies of a family of schemata describe the multivariate marginal distribution of the population over the defined locii of the schemata. Thus another way to state our contribution is that we give sufficient conditions under which the multivariate marginal distribution of an evolving population over a small number of locii can be numerically approximated over multiple generations regardless of the length of the genomes.

We stress that our use of the term `coarse-graining' differs from the way this term has been used in other publications. For instance in \cite{Stephens:2003:gecco} the term `coarse-graining' is used to describe a reduction of the IPGA equations such that each equation in the new system is similar in form to the equations in the original system. The state variables in the new system are defined in terms of the state variables in the original system. Therefore a numerical iteration of the the new system is only computationally tractable when the length of the genomes is relatively short. Elsewhere the term coarse-graining has been defined as ``a collection of subsets of the search space that covers the search space''\cite{conf/gecco/ContrerasRS03}, and as ``just a function from a genotype set to some other set"\cite{interPopConstaints}.

\subsection{Some Previous Coarse-Graining Results}

Techniques from statistical mechanics have been used to coarse-grain GA dynamics in \cite{statmech1994,shapiro1997,shap2001}  (see \cite{shapiro:2001:smtga} for a survey of applications of statistical mechanics approaches to GAs). The macroscopic variables of these coarse-grainings are the first few cumulants of the fitness distribution of the evolving population. In \cite{journals/tcs/RoweVW06} several exact coarse-graining results are derived for an IPGA whose variation operation is limited to mutation.

Wright et. al. show in \cite{conf/gecco/WrightVR03} that the dynamics of a non-selective IPGA  can be coarse-grained such that the macroscopic variables are the frequencies of a family of schemata in a schema partition. However they argue that the dynamics of a regular selecto-mutato-recombinative IPGA  cannot be similarly coarse-grained ``except in the trivial case where fitness is a constant for each schema in a schema family"\cite{conf/gecco/WrightVR03}. Let us call this condition \emph{schematic fitness invariance}. Wright et. al. imply that it is so severe that it renders the coarse-graining result essentially useless.

This negative result holds true when there is no constraint on the initial population. In this paper we show that if we constrain the class of initial populations then it is possible to coarse-grain the dynamics of a  regular IPGA under a much \emph{weaker} constraint on the fitness function. The constraint on the class of initial populations is not onerous; this class includes the uniform distribution over the genome set.

\subsection{Structure of this Paper}

The rest of this paper is organized as follows:  in the next section we define the basic mathematical objects and notation which we use to model the dynamics of an infinite population \emph{evolutionary} algorithm (IPEA). This framework is very general; we make no commitment to the data-structure of the genomes, the nature of mutation, the nature of recombination , or the number of parents involved in a recombination. We do however require that selection be fitness proportional.  In section 3 we define the concepts of semi-coarsenablity, coarsenablity and global coarsenablity which allow us to formalize a useful class of exact coarse-grainings. In section 4 and section 5 we prove some stepping-stone results about selection and variation. We use these results in section 6 where we prove that an IPEA that satisfies certain abstract conditions can be coarse-grained. The proofs in sections 5 and 6 rely on lemmas which have been relegated to and proved in the appendix. In section 7 we specify concrete conditions under which IPGAs with long genomes and non-trivial fitness functions can be coarse-grained such that the macroscopic variables are schema frequencies and the fidelity of the coarse-graining is likely to be high. We conclude in section 8 with a summary of our work.

\section{\large Mathematical Preliminaries}
\label{MathematicalPreliminaries}
Let $X,Y$ be sets and let $\xi:X\rightarrow Y$ be some function. For any $y\in Y$ we use the notation
$\langle y\rangle_{\!\xi}^{\phantom g} $ to denote the pre-image of $y$, i.e. the set
$\{x\in X\,|\,\beta(x)=y\}$. For any subset $A\subset X$ we use the notation $\xi(A)$ to denote the set $\{y\in Y|\,\xi(a)=y \text{ and } a\in A\}$

As in \cite{ToussaintThesis}, for any set $X$ we use the notation $\Lambda^X$ to denote
the set of all distributions over $X$, i.e. $\Lambda^X$ denotes set $\{f:X\rightarrow
[0,1] \,\,|\,\, \sum_{x\in X} f(x)=1\}$. For any set $X$, let $0^X:X\rightarrow\{0\}$ be
the constant zero function over $X$. For any set $X$, an $m$-parent
transmission function \cite{MontgomerySlatkin05011970,kinnear:altenberg,toussaint:03-foga} over $X$  is an element of the set
\[\bigg\{T:\prod_1^{m+1} X\rightarrow[0,1]\,\,\bigg|\,\,\forall x_1,
\ldots,x_m\in  X, \sum_{x\in X}T(x,x_1',\ldots, x_m')=1\bigg\}\]

Extending the notation introduced above, we denote this set by $\Lambda^X_{m}$. Following \cite{ToussaintThesis}, we use conditional probability notation in our denotation of transmission functions. Thus an $m$-parent transmission function $T(x,x_1,\ldots,x_{m})$ is denoted $T(x|x_{1},\ldots,x_{m})$.

A transmission function can be used to model the individual-level effect of mutation, which operates on one parent and produces one child, and indeed the individual-level effect of any variation operation which operates on any numbers of parents and produces one child.

Our scheme for modeling EA dynamics is based on the one used in \cite{ToussaintThesis}. We model the genomic populations of an EA as distributions over the genome set. The population-level effect of the evolutionary operations of an EA is modeled by mathematical operators whose inputs and outputs are such distributions.

The expectation operator, defined below, is used in the definition of the selection operator, which follows thereafter.

\begin{definition}\,\,\textsc{(Expectation Operator)}\label{WeightedAverageOperator}
Let $X$ be some finite set, and let $f:X\rightarrow \mathbb R^+$ be some function. We define the expectation operator $\mathcal E_f:\Lambda^X\cup 0^X \rightarrow \mathbb
R^+\cup\{0\}$ as follows:
\[\mathcal E_f(p) =\sum\limits_{x\in X}f(x)p(x)\]
\end{definition}

The selection operator is parameterized by a fitness function. It models the effect of fitness proportional selection on a population of genomes.
\begin{definition}\,\,\textsc{(Selection Operator)}\label{SelectionOperator}
Let $X$ be some finite set and let $f:X\rightarrow \mathbb R^+$ be some function. We
define the \emph{Selection Operator} $\mathcal S_f:\Lambda^X \rightarrow \Lambda^X$ as
follows: \[(\mathcal
S_fp)(x)=\frac{f(x)p(x)}{\mathcal E_f(p)}\]
\end{definition}

The population-level effect of variation is modeled by the variation operator. This operator is parameterized by a transmission function which models the effect of variation at the individual level.

\begin{definition}\,\,\textsc{(Variation Operator\footnote{also called the Mixing Operator
in \cite{vose:1999:sgaft} and \cite{ToussaintThesis}})}\label{Transmission functionOperator} Let
$X$ be a countable set, and for any $m\in \mathbb N^+$, let $T\in \Lambda^X_{m}$ be a transmission
function over $X$. We define the variation operator $\mathcal V^{\phantom |}_T:\Lambda^X
\rightarrow \Lambda^X$ as follows: \[(\mathcal V^{\phantom |}_Tp)(x) =\sum_{\substack{(x_1,\ldots,x_m)\\\in\,\prod_1^mX}}
T(x|x_1,\ldots,x_m)\prod_{i=1}^mp(x_i)\]
\end{definition}

The next definition describes the projection operator (previously used in
\cite{vose:1999:sgaft} and \cite{ToussaintThesis}). A projection operator that is parameterized by some function $\beta$ `projects' distributions over the domain of $\beta$, to distributions over its co-domain.

\begin{definition}\,\,\textsc{(Projection Operator)} \label{ProjectionOperator}
Let $X$ be a countable set, let $Y$ be some set, and let $\beta:X\rightarrow Y$ be a function. We define the
projection operator, $\Xi_\beta:\Lambda^X\rightarrow \Lambda^Y$ as follows:
\[(\Xi_\beta \p)(y)=\sum_{x\in\langle y\rangle_{\!\beta}^{\phantom g}}p(x)\]
and call $\Xi_\beta \p$ the $\beta$-projection of $p$.
\end{definition}

\section{Formalization of a Class of Coarse-Grainings}

The following definition introduces some convenient function-related terminology.
\begin{definition}\,\,\textsc{(Partitioning, Theme Set, Themes, Theme Class)}\label{ThemeThemeSetThemeClass} Let $X$, $K$ be sets and let
$\beta:X\rightarrow K$ be a surjective function. We call $\beta$ a \emph{partitioning}, call the co-domain $K$ of $\beta$ the \emph{theme
set} of $\beta$, call any element in $K$ a \emph{theme} of $\beta$, and call the pre-image $\langle k
\rangle_{\!\beta}^{\phantom g}$ of some $k\in K$, the \emph{theme class} of $k$ under $\beta$.
\end{definition}

The next definition formalizes a class of coarse-grainings in which the macroscopic and microscopic state variables always sum to 1.
\begin{definition}[Semi-Coarsenablity, Coarsenablity, Global Coarsenablity] Let $G, K$ be sets, let $\mathcal W:\Lambda^G\rightarrow\Lambda^G$ be an operator, let $\beta:G\rightarrow K$ be a partitioning, and let $U\subseteq\Lambda^G$ such that $\Xi_\beta(U)=\Lambda^K$. We say that $\mathcal W$ is semi-coarsenable under $\beta$ on $U$ if there exists an operator $\mathcal Q:\Lambda^K\rightarrow \Lambda^K$ such that for all
$p\in U$, $\mathcal Q\circ\Xi_\beta p=\Xi_\beta\circ \mathcal Wp$,
i.e. the following diagram commutes:
\[\xymatrix{
 U \ar[rr]^{\mathcal W} \ar[d]_{\Xi_\beta} && {\Lambda^G}
 \ar[d]^{\Xi_\beta}\\
    \Lambda^K\ar[rr]_{\mathcal Q} && \Lambda^K}\]
Since $\beta$ is surjective, if $\mathcal Q$ exists, it is clearly unique; we call it the quotient. We call $ G, K, W, \text{ and }U$ the domain, co-domain, primary operator and turf respectively. If in addition $\mathcal W(U)\subseteq U$ we say that $\mathcal W$ is coarsenable under $\beta$ on $U$. If in addition $U=\Lambda^G$ we say that $\mathcal W$ is globally coarsenable under $\beta$.
\end{definition}
Note that the partition function $\Xi_\beta$ of the coarse-graining is not the same as the partitioning $\beta$ of the coarsening.

Global coarsenablity is a stricter condition than coarsenablity, which in turn is a stricter condition than semi-coarsenablity. It is easily shown that global coarsenablity is equivalent to Vose's notion of compatibility \cite[p. 188]{vose:1999:sgaft} (for a proof see Theorem 17.5 in \cite{vose:1999:sgaft}).

If some operator $\mathcal W$ is coarsenable under some function $\beta$ on some turf $U$ with some quotient $\mathcal Q$, then for any distribution $\p_K\in\Xi_\beta(U)$, and all distributions $\p_G\in\langle \p_K\ranglel_{\Xi_\beta}$, one can study the \emph{projected} effect of the repeated application of $\mathcal W$ to $\p_G$ simply by studying the effect of the repeated application of $\mathcal Q$ to $\p_K$. If the size of $K$ is small then a computational study of the projected effect of the repeated application of $\mathcal W$ to distributions in $U$ becomes feasible.

\section{Global Coarsenablity of Variation}\label{SelectionalConstraints}
We show that some variation operator $\mathcal V_T$ is globally coarsenable under some partitioning if a relationship, that we call \emph{ambivalence}, exists between the transmission function $T$ of the variation operator and the partitioning.

To illustrate the idea of ambivalence consider a partitioning  $\beta$ which partitions a genome set $G$ into three subsets. Fig 1 depicts the behavior of a two-parent transmission function that is ambivalent under $\beta$. Given two parents and some child, the probability that the child will belong to some theme class depends \emph{only} on the theme classes of the parents and \emph{not} on the specific parent genomes. Hence the name `ambivalent' --- it captures the sense that when viewed from the coarse-grained level of the theme classes, a transmission function `does not care' about the specific genomes of the parents or the child.

\begin{figure}[t]\label{ambivalencfigure}\begin{center}
\includegraphics[height=4.5cm, width=7.8cm]{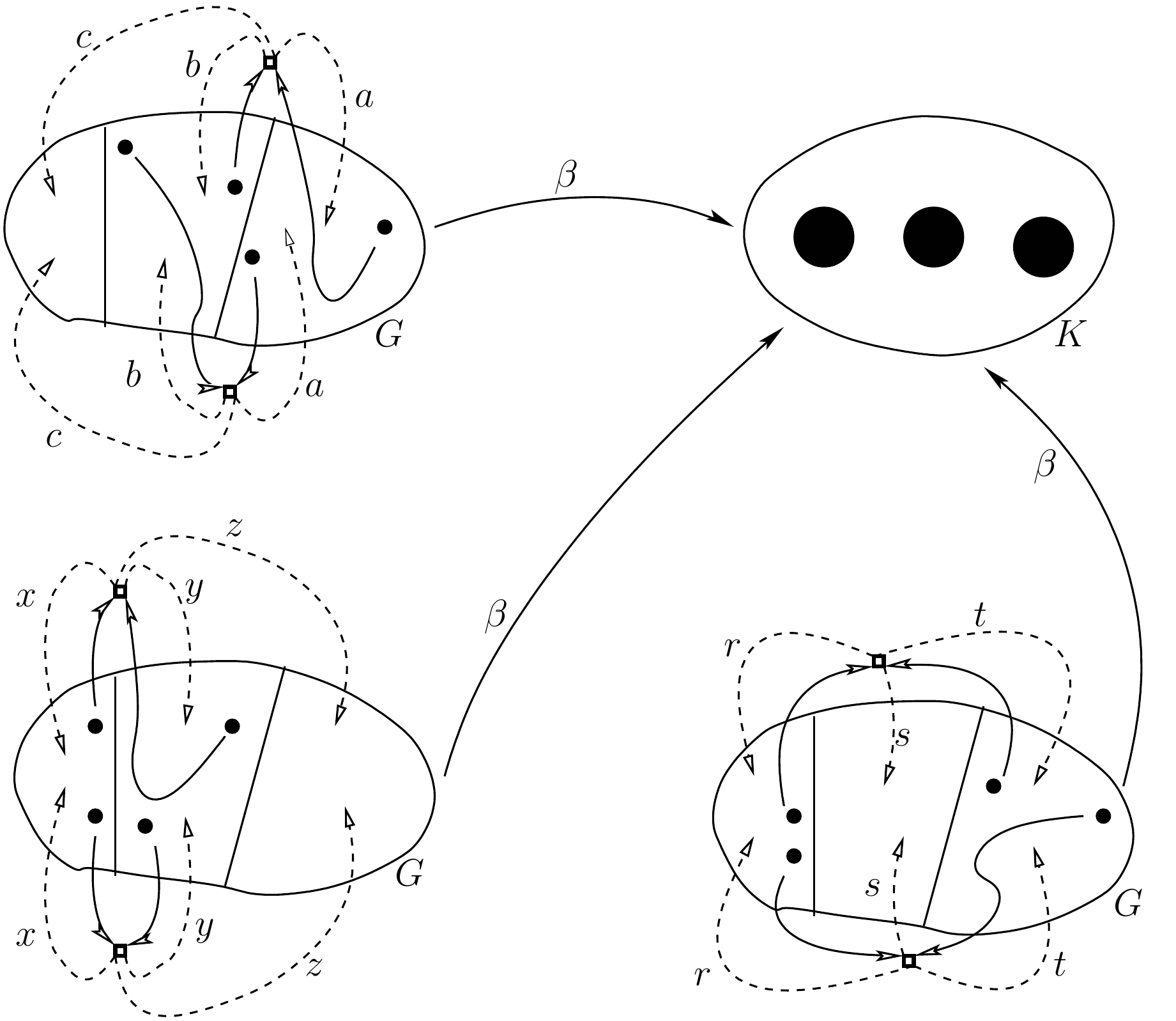}\end{center}
\caption{small Let $\beta:G\rightarrow K$ be a coarse-graining which partitions the genome
set $G$ into three theme classes. This figure depicts the behavior of a two-parent
variation operator that is ambivalent under $\beta$. The small dots denote specific
genomes and the solid unlabeled arrows denote the recombination of these genomes. A dashed
arrow denotes that a child from a recombination may be produced `somewhere' within the
theme class that it points to, and the label of a dashed arrow denotes the probability
with which this might occur. As the diagram shows the probability that the child of a
variation operation will belong to a particular theme class depends \emph{only} on the
theme classes of the parents and \emph{not} on their specific genomes}
\end{figure}
\normalsize
The definition of ambivalence that follows is equivalent to but more useful than the definition given in \cite{interPopConstaints}
\begin{definition}\,\,\textsc{(Ambivalence)} Let $G, K$ be countable sets, let  $T\in
\Lambda^G_{m}$ be a transmission function, and let $\beta:G\rightarrow K$ be a partitioning. We say that $T$ is ambivalent under $\beta$ if there exists some transmission function $D\in
\Lambda^K_{m}$, such that for all $k,k_1,\ldots, k_m\in K$ and for any $x_1\in\langle
k_1\rangle_{\!\beta}^{\phantom g},\ldots,x_m\in\langle k_m\rangle_{\!\beta}^{\phantom
g}$,
\[\sum_{x\in \langle k
\rangle_{\!\beta}^{\phantom g}} T(x|x_1,\ldots,x_m)=D(k|k_1,\ldots, k_m) \] If such a $D$ exits, it is clearly unique. We denote it by $
T^{\overrightarrow\beta}$ and call it the theme transmission function.
\end{definition}

\vspace{-.2cm}Suppose $T\in \Lambda^X_{m}$ is ambivalent under some
$\beta:X\rightarrow K$, we can use the projection operator to express the projection of $T$ under $\beta$ as follows: for all $k, k_1,\ldots, k_m\in K$, and any $x_1\in\langle
k_1\rangle_{\!\beta}^{\phantom g}, \ldots,x_m\in\langle k_m\rangle_{\!\beta}^{\phantom
g}$, $T^{\overrightarrow\beta}(k|k_1,\ldots k_m)$ is given by $(\Xi_\beta
(T(\cdot\,|x_1,\ldots,x_m)))(k)$. The notion of ambivalence is equivalent to a generalization of Toussaint's notion of trivial neutrality \cite[p. 26]{ToussaintThesis}. A one-parent   transmission function is ambivalent under a mapping to the set of phenotypes if and only if it is trivially neutral.

The following theorem shows that a variation operator is globally coarsenable under some partitioning if it is parameterized by a transmission function which is ambivalent under that partitioning\ifthenelse{\boolean{showProofs}}{}{\footnote{Due to space
restrictions proofs have been omitted. They can be found in the full version of this
paper which is posted on the author's website}}. The method by which we prove this theorem extends the method used in the proof of Theorem 1.2.2 in \cite{ToussaintThesis}.

\begin{theorem}[Global Coarsenablity of Variation] \label{globConcOfVar} Let $G$ and $K$ be countable sets, let
$T\in \Lambda^G_{m}$ be a transmission function and let  $\beta:G\rightarrow K$ be some
partitioning such that $T$ is ambivalent under $\beta$. Then $\mathcal V_T:\Lambda^G\rightarrow \Lambda^G$ is globally coarsenable under $\beta$ with quotient $\mathcal V_{T^{\overrightarrow\beta}}$, i.e. the following diagram commutes:
\nopagebreak
\[\xymatrix{
 \Lambda^G \ar[rr]^{\mathcal V_T} \ar[d]_{\Xi_\beta} && {\Lambda^G}
 \ar[d]^{\Xi_\beta}\\
    \Lambda^K\ar[rr]_{\mathcal V_{T^{\overrightarrow\beta}}} && \Lambda^K}\]
\end{theorem}
\ifthenelse{\boolean{showProofs}}{
\textsc{Proof:  } For any $p\in\Lambda^G$,
{\allowdisplaybreaks
\begin{align*}
&(\Xi_\beta\circ\mathcal V_Tp)(k)\\ &=\sum_{x\in \langle k\rangle_{\!\beta}^{\phantom
g}}
\sum_{\substack{(x_1,\ldots,x_m)\\\in\prod\limits_1^{m}X}}T(x|x_1,\ldots,x_m)\prod_{i=1}^mp(x_i)\\
&=\sum_{\substack{(x_1,\ldots,x_m)\\\in\prod\limits_1^{m}X}}
\sum_{x\in \langle k\rangle_{\!\beta}^{\phantom g}}T(x|x_1,\ldots,x_m)\prod_{i=1}^mp(x_i)\\
&=\sum_{\substack{(x_1,\ldots,x_m)\\\in\prod\limits_1^{m}X}}\prod_{i=1}^mp(x_i)
\sum_{x\in \langle k\rangle_{\!\beta}^{\phantom g}}T(x|x_1,\ldots,x_m)\\
&=\sum_{\substack{(k_1,\ldots,k_m)\\\in\prod\limits_1^{m}K}}\sum_{\substack{(x_1,\ldots,x_m)\\\in\prod\limits_{j=1}^{m}\langle
k_j\rangle_{\!\beta}^{\phantom g}}}\prod_{i=1}^mp(x_i) \sum_{x\in \langle k\rangle_{\!\beta}^{\phantom g}}T(x|x_1,\ldots,x_m)\\
&=\sum_{\substack{(k_1,\ldots,k_m)\\\in\prod\limits_1^{m}K}}\sum_{\substack{(x_1,\ldots,x_m)\\\in\prod\limits_{j=1}^{m}\langle
k_j\rangle_{\!\beta}^{\phantom g}}}\prod_{i=1}^mp(x_i)
T^{\overrightarrow \beta}(k|k_1,\ldots,k_m)\\
&=\sum_{\substack{(k_1,\ldots,k_m)\\\in\prod\limits_1^{m}K}}T^{\overrightarrow
\beta}(k|k_1,\ldots,k_m)\sum_{\substack{(x_1,\ldots,x_m)\\\in\prod\limits_{j=1}^{m}\langle
k_j\rangle_{\!\beta}^{\phantom g}}}\prod_{i=1}^mp(x_i)\\
&=\sum_{\substack{(k_1,\ldots,k_m)\\\in\prod\limits_1^{m}K}}T^{\overrightarrow
\beta}(k|k_1,\ldots,k_m)\sum_{x_1\in\langle
k_1\rangle_{\!\beta}^{\phantom g}}\ldots\sum_{x_m\in\langle
k_m\rangle_{\!\beta}^{\phantom g}}p(x_1)\ldots p(x_m)\\
&=\sum_{\substack{(k_1,\ldots,k_m)
\\\in\prod\limits_1^{m}K}}T^{\overrightarrow
\beta}(k|k_1,\ldots,k_m)\bigg(\sum_{x_1\in\langle
k_1\rangle}p(x_1)\bigg)\ldots\bigg(\sum_{x_m\in\langle
k_m\rangle}p(x_m)\bigg)\\
&=\sum_{\substack{(k_1,\ldots,k_m)\\\in\prod\limits_1^{m}K}}T^{\overrightarrow
\beta}(k|k_1,\ldots,k_m)\prod_{i=1}^m\bigg((\Xi_\beta p)(k_i)\bigg)\\ &=(\mathcal
V_{T^{\overrightarrow
\beta}}\circ\Xi_\beta p)(k)\quad\quad \qed
\end{align*}
}}{}

The implicit parallelism theorem in \cite{conf/gecco/WrightVR03} is similar to the theorem above. Note however that the former theorem only shows that variation is globally coarsenable if firstly, the genome set consists of ``fixed length strings, where the size of the alphabet can vary from position to position", secondly the partition over the genome set is a schema partition, and thirdly variation is `structural' (see \cite{conf/gecco/WrightVR03} for details). The global coarsenablity of variation theorem has none of these specific requirements. Instead it is premised on the existence of an abstract relationship -- ambivalence --  between the variation operation and a partitioning. The abstract nature of this relationship makes this theorem applicable to evolutionary algorithms other than GAs. In addition this theorem illuminates the essential relationship between `structural' variation and schemata which was used (implicitly) in the proof of the implicit parallelism theorem.

In \cite{interPopConstaints} it is shown that a variation operator that models any  combination of variation operations that are commonly used in GAs --- i.e. any combination of mask based crossover and `canonical' mutation, in any order --- is  ambivalent under any partitioning that maps bitstrings to schemata (such a partitioning is called a schema partitioning). Therefore `common' variation in IPGAs is globally coarsenable under \emph{any} schema partitioning. This is precisely the result of the implicit parallelism theorem.

\section{Limitwise Semi-Coarsenablity of Selection}
For some fitness function $f:G\rightarrow \mathbb R^+$ and some partitioning  $\beta:G\rightarrow K$ let us say that $f$ is \emph{thematically invariant} under $\beta$ if, for any schema $k\in K$, the genomes that belong to $\langle k \rangle_\beta$ all have the same fitness.
Paraphrasing the discussion in \cite{conf/gecco/WrightVR03} using the terminology developed in this paper, Wright et. al. argue that if the selection operator  is globally coarsenable under some schema partitioning $\beta:G\rightarrow K$ then the fitness function that parameterizes the selection operator is `schematically' invariant under $\beta$. It is relatively simple to use contradiction to prove a generalization of this statement for arbitrary partitionings.

Schematic invariance is a very strict condition for a fitness function. An IPGA whose fitness function meets this condition is unlikely to yield any substantive information about the dynamics of real world GAs.

As stated above, the selection operator is not \emph{globally} coarsenable unless the fitness function satisfies thematic invariance, however if the set of distributions that selection operates over (i.e. the turf) is appropriately constrained, then, as we show in this section, the selection operator is \emph{semi-}coarsenable over the turf even when the fitness function only satisfies a much \emph{weaker} condition called thematic \emph{mean} invariance.\\

For any partitioning $\beta:G\rightarrow K$, any theme $k$, and any distribution $p\in\Lambda^G$, the theme conditional operator, defined below, returns a conditional distribution in  $\Lambda^G$ that is obtained by normalizing the probability mass of the elements in $\langle k\rangle_\beta$ by $(\Xi_\beta p)(k)$

\begin{definition}[Theme Conditional Operator] Let $G$ be some countable set, let $K$ be some set, and let $\beta:G\rightarrow K$ be some function. We define the theme conditional operator $\mathcal C_\beta:\Lambda^G\times K\rightarrow \Lambda^G\cup 0^G$ as follow: For any $p\in\Lambda^G$, and any $k\in K$, $\mathcal C_\beta(p,k)\in\Lambda^G\cup 0^G$ such that for any $x\in \langle k \rangle_\beta$,
\[(\mathcal C_\beta(p,k))(x)=\left\{\begin{array}{cl}0&\text{if } (\Xi_\beta p)(k)=0\\\frac{p(x)}{(\Xi_\beta p)(k)}& \text{otherwise}\end{array}\right.\]
\end{definition}

A useful property of the theme conditional operator is that it can be composed with the expected fitness operator to give an operator that returns the average fitness of the genomes in some theme class. To be precise, given some finite genome set $G$, some partitioning $\beta:G\rightarrow K$, some fitness function $f:G\rightarrow \mathbb R^+$, some distribution $p\in \Lambda^G$, and some theme $k\in K$, $\mathcal E_f\circ\mathcal C_\beta(p,k)$ is the average fitness of the genomes in $\langle k \rangle_\beta$. This property proves useful in the following definition.

\begin{definition}[Bounded Thematic Mean Divergence, Thematic Mean Invariance] Let $G$ be some finite set, let $K$ be some set, let $\beta:G\rightarrow K$ be a partitioning, let $f:G\rightarrow\mathbb R^+$ and $f^*:K\rightarrow \mathbb R^+$ be functions, let $U\subseteq\Lambda^G$, and let $\delta\in \mathbb R_0^+$. We say that the thematic mean divergence of $f$ with respect to $f^*$ on $U$ under $\beta$ is bounded by $\delta$ if, for any $p\in U$ and for any $k\in K$
\[|\mathcal E_f\circ\mathcal C_\beta(p, k)-f^*(k)|\leq\delta\]
If $\delta=0$ we say that $f$ is thematically mean invariant with respect to $f^*$ on $U$
\end{definition}

The next definition gives us a means to measure a `distance' between real valued functions over finite sets.
\begin{definition}[Manhattan Distance Between Real Valued Functions] Let $X$ be a finite set then for any functions $f,h$ of type $X\rightarrow \mathbb R$ we define the manhattan distance between $f$ and $h$, denoted by $d(f,h)$, as follows:
\[d(f,h)=\sum_{x\in X}|f(x)-h(x)|\]
\end{definition}
\noindent It is easily checked that $d$ is a metric.

Let $f:G\rightarrow \mathbb R^+$, $\beta:G\rightarrow K$ and $f^*:K\rightarrow \mathbb R^+$ be functions with finite domains, and let $U\in \Lambda^G$. The following theorem shows that if the thematic mean divergence of $f$ with respect to $f^*$ on $U$ under $\beta$ is bounded by some $\delta$, then in the limit as $\delta\rightarrow 0$, $\mathcal S_f$ is semi-coarsenable under $\beta$ on  $U$ .

\begin{theorem}[Limitwise Semi-Coarsenablity of Selection]\label{SelectionUnderCoarsegraining}
Let $G$ and $K$ be finite sets, let $\beta:G\rightarrow K$ be a partitioning,  Let $U\subseteq\Lambda^G$ such that $\Xi_\beta(U)=\Lambda^K$, let $f:G\rightarrow \mathbb R^+$, $f^*:K\rightarrow \mathbb R^+$ be some functions such that the thematic mean divergence of $f$ with respect to $f^*$ on $U$ under $\beta$ is bounded by $\delta$, then for any $p\in U$ and any $\epsilon>0$ there exists a $\delta'>0$ such that,  \[\delta<\delta'\Rightarrow d(\Xi_\beta \circ \mathcal S_fp, \mathcal S_{f^*}\circ \Xi_\beta p)<\epsilon\]
\end{theorem}
\noindent We depict the result of this theorem as follows:
\[\xymatrix{
 U \ar[rr]^{\mathcal S_f} \ar[d]_{\Xi_\beta} \ar @{}[rrd] | {\lim\limits_{\delta\rightarrow 0}}&& {\Lambda^G}
 \ar[d]^{\Xi_\beta} \\
    \Lambda^K\ar[rr]_{\mathcal S_{f^*}} && \Lambda^K}\]

\ifthenelse{\boolean{showProofs}}{
\noindent \textsc{Proof:  } For any $p\in U$ and for any $k \in K$,
{\allowdisplaybreaks
\begin{align*}
&(\Xi_\beta\circ\mathcal S_{f}p)(k)\\&=\sum_{g\in\langle k
\rangle_{\!\beta}^{\phantom g}}(\mathcal S_{f}p)(g)\\
&=\sum_{g\in\langle k\rangle_{\!\beta}^{\phantom g}}\frac{f(g).p(g)}{\sum_{g'\in G}f(g').p(g')}
\\&=\frac{\sum\limits_{g\in\langle
k\rangle_{\!\beta}^{\phantom g}}f(g).(\Xi_\beta p)(k).
(\mathcal C_\beta(p,k))(g)}{\sum\limits_{k'\in K}\sum\limits_{g'\in
\langle k' \rangle_{\!\beta}^{\phantom g}} f(g').(\Xi_\beta p)(k')(\mathcal C_\beta(p,k'))(g')} \\
&=\frac{(\Xi_\beta p)(k)\sum\limits_{g\in\langle
k\rangle_{\!\beta}^{\phantom g}}f(g).(\mathcal C_\beta(p,k))(g)}{\sum\limits_{k'\in K}(\Xi_\beta p)(k')\sum\limits_{g'\in \langle
k'\rangle_{\!\beta}^{\phantom g}}
f(g').(\mathcal C_\beta(p,k'))(g')}\\
&=\frac{(\Xi_\beta p)(k).\mathcal E_f\circ\mathcal C_\beta(p,k)}{\sum\limits_{k'\in K}(\Xi_\beta p^{\phantom
t}_G)(k').\mathcal E_f\circ\mathcal C_\beta(p,k')}\\
&=(\mathcal S_{\mathcal E_f\circ\mathcal C_\beta(p,\cdot)}\circ\Xi_\beta p)(k)
\end{align*}
So we have that \[ d(\Xi_\beta \circ\mathcal S_fp, \mathcal S_{f^*}\circ\Xi_\beta p) = d(\mathcal S_{\mathcal E_f\circ\mathcal C_\beta(p,\cdot)}\circ\Xi_\beta p, \mathcal S_{f^*}\circ\Xi_\beta p)\]

\noindent By Lemma, \ref{lemmaD} (in the appendix) for any $\epsilon>0$ there exists a $\delta_1>0$ such that,
\[d(\mathcal E_f\circ\mathcal C_\beta(p,.),f^*)<\delta_1\Rightarrow d(\mathcal S_{\mathcal E_f\circ\mathcal C_\beta(p,\cdot)}(\Xi_\beta p), \mathcal S_{f^*}(\Xi_\beta p))<\epsilon\]

\noindent Now, if $\delta<\frac{\delta'}{|K|}$, then $d(\mathcal E_f\circ\mathcal C_\beta(p,.),f^*)<\delta_1  \quad\quad \qed$
}}{}
\\
\begin{corollary} If $\delta=0$, i.e. if $f$ is thematically mean invariant with respect to $f^*$ on $U$, then $S_f$ is semi-coarsenable under $\beta$ on $U$ with quotient $\mathcal S_{f^*}$, i.e. the following diagram commutes:
\[\xymatrix{
 U \ar[rr]^{\mathcal S_f} \ar[d]_{\Xi_\beta} && {\Lambda^G}
 \ar[d]^{\Xi_\beta} \\
    \Lambda^K\ar[rr]_{\mathcal S_{f^*}} && \Lambda^K}\]\end{corollary}

\section{Limitwise Coarsenablity of Evolution}

The two definitions below formalize the idea of an infinite population model of an EA, and its dynamics \footnote{The definition of an EM given here is different from its definition in \cite{Burjorjee05-ThRepWkshp,conf/iicai/BurjorjeeP05}. The fitness function in this definition maps genomes directly to fitness values. It therefore subsumes the genotype-to-phenotype and the phenotype-to-fitness functions of the previous definition. In previous work these two functions were always composed together; their subsumption within a single function increases clarity.}.

\begin{definition}[Evolution Machine] An evolution machine (EM) is a tuple $(G, T, f)$ where $G$ is some set called the domain, $f:G\rightarrow \mathbb R^+$ is a function called the fitness function and $T\in \Lambda^G_m$ is called the transmission function.
\end{definition}

\begin{definition}[Evolution Epoch Operator] Let $E=(G,T,f)$ be an evolution machine. We define the evolution epoch operator $\mathcal G_E:\Lambda^G\rightarrow \Lambda^G$ as follows:
\[\mathcal G_E=\mathcal V_T\circ\mathcal S_f\]\end{definition}

For some evolution machine $E$, our aim is to give sufficient conditions under which, for any $t\in \mathbb Z^+$, $\mathcal G^t_E$ approaches coarsenablity in the limit. The following definition gives us a formal way to state one of these conditions.

\begin{definition}[Non-Departure] Let $E=(G,T,f)$ be an evolution machine, and let $U\subseteq\Lambda^G$. We say that $E$ is \emph{non-departing} over $U$ if
\[\mathcal V_T\circ \mathcal S_f(U)\subseteq U\]
\end{definition}

\noindent Note that our definition does \emph{not} require $S_f(U)\subseteq U$ in order for $E$ to be non-departing over $U$.

\begin{theorem}[Limitwise Coarsenablity of Evolution]
Let $E=(G,T,f)$, be an evolution machine such that $G$ is finite, let $\beta:G\rightarrow K$ be some partitioning, let $f^*:K\rightarrow \mathbb R^+$ be some function, let $\delta\in\mathbb R^+_0$, and let $U\subseteq\Lambda^G$ such that $\Xi_\beta(U)=\Lambda^K$. Suppose that the following statements are true:
\begin{enumerate}
\item The thematic mean divergence of $f$ with respect to $f^*$ on $U$ under $\beta$ is bounded by $\delta$
\item $T$ is ambivalent under $\beta$
\item $E$ is non-departing over $U$
\end{enumerate}
Then, letting $E^*=(K, T^{\overrightarrow{\beta}}, f^*)$ be an evolution machine, for any $t\in \mathbb Z^+$ and any $p\in U$,
\begin{enumerate}
\item $\mathcal G^t_Ep\,\,\in U$
\item For any $\epsilon>0$, there exists $\delta'>0$ such that,  \[\delta<\delta' \Rightarrow d(\Xi_\beta\circ \mathcal G^t_Ep\,\,,\,\, \mathcal G^t_{E^*}\circ\Xi_\beta p)<\epsilon\]\end{enumerate}
\end{theorem}

\noindent We depict the result of this theorem as follows:
\[\xymatrix{
 U \ar[rr]^{\mathcal G^t_E} \ar[d]_{\Xi_\beta} \ar @{}[rrd] | {\lim\limits_{\delta\rightarrow 0}}&& U
 \ar[d]^{\Xi_\beta}\\
    \Lambda^K\ar[rr]_{\mathcal G^t_{E^*}} && \Lambda^K}\]
\ifthenelse{\boolean{showProofs}}{
{\sc Proof:  }
We prove the theorem for any $t\in \mathbb Z^+_0$. The proof is by induction on $t$. The base case, when $t=0$, is trivial. For some $n=\mathbb Z^+_0$, let us assume the hypothesis for $t=n$. We now show that it is true for $t=n+1$. For any $p\in U$,  by the inductive assumption $\mathcal G_E^np$ is in $U$. Therefore, since $E$ is non-departing over $U$, $\mathcal G_E^{n+1}p\,\,\in U$. This completes the proof of the first part of the hypothesis. For a proof of the second part note that,
\begin{align*}
&d(\Xi_\beta\circ\mathcal G^{n+1}_Ep\, , \, \mathcal G^{n+1}_{E^*}\circ\Xi_\beta\p)\\
&=d(\Xi_\beta\circ\mathcal V_T\circ\mathcal S_f\circ\mathcal G^{n}_Ep\, , \, \mathcal V_{T^{\overrightarrow\beta}}\circ\mathcal S_{f^*}\circ\mathcal G^{n}_{E^*}\circ\Xi_\beta\p)\\
&=d(\mathcal V_{T^{\overrightarrow\beta}}\circ\Xi_\beta\circ\mathcal S_f\circ\mathcal G^{n}_Ep\, , \, \mathcal V_{T^{\overrightarrow\beta}}\circ\mathcal S_{f^*}\circ\mathcal G^{n}_{E^*}\circ\Xi_\beta p)&\text{(by theorem \ref{globConcOfVar})}
\end{align*}

\noindent Hence, for any $\epsilon>0$, by Lemma \ref{lemmaB} there exists $\delta_1$ such that
\begin{align*}d(\Xi_\beta\circ\mathcal S_f\circ\mathcal G^{n}_Ep\, , \, \mathcal \mathcal S_{f^*}\circ\mathcal G^{n}_{E^*}\circ\Xi_\beta p)<\delta_1\Rightarrow  d(\Xi_\beta\circ\mathcal G^{n+1}_Ep\, , \, \mathcal G^{n+1}_{E^*}\circ\Xi_\beta\p)<\epsilon\end{align*}
\noindent As $d$ is a metric it satisfies the triangle inequality. Therefore we have that
\begin{multline*}
d(\Xi_\beta\circ\mathcal S_f\circ\mathcal G^{n}_Ep\, , \, \mathcal S_{f^*}\circ\mathcal G^{n}_{E^*}\circ\Xi_\beta p)\leq\\
d(\Xi_\beta\circ\mathcal S_{f}\circ \mathcal G^{n}_Ep\, , \, \mathcal S_{f^*}\circ\Xi_\beta\circ\mathcal G^{n}_{E} p)+\\d(\mathcal S_{f^*}\circ\Xi_\beta\circ\mathcal G^{n}_{E} p\, , \, \mathcal S_{f^*}\circ\mathcal G^{n}_{E^*}\circ\Xi_\beta p)
\end{multline*}
By our inductive assumption $\mathcal G_E^np\,\,\in U$. So, by  theorem \ref{SelectionUnderCoarsegraining} there exists a $\delta_2$ such that
\[\delta<\delta_2\Rightarrow d(\Xi_\beta\circ\mathcal S_{f}\circ \mathcal G^{n}_Ep\, , \, \mathcal S_{f^*}\circ\Xi_\beta\circ\mathcal G^{n}_{E} p)<\frac{\delta_1}{2}\]

\noindent By lemma \ref{lemmaC} there exists a $\delta_3$ such that
\begin{multline*}d(\Xi_\beta\circ\mathcal G^{n}_{E} p\, , \, \mathcal G^{n}_{E^*}\circ\Xi_\beta p)<\delta_3\Rightarrow d(\mathcal S_{f^*}\circ\Xi_\beta\circ\mathcal G^{n}_{E} p\, , \, \mathcal S_{f^*}\circ\mathcal G^{n}_{E^*}\circ\Xi_\beta p)<\frac{\delta_1}{2}\end{multline*}

\noindent By our inductive assumption, there exists a $\delta_4$ such that \[\delta<\delta_4 \Rightarrow  d(\Xi_\beta\circ\mathcal G^{n}_{E} p\, , \, \mathcal G^{n}_{E^*}\circ\Xi_\beta p)<\delta_3\]

\noindent Therefore, letting $\delta'=\text{min}(\delta_2, \delta_4)$ we get that
\[\delta<\delta^*\Rightarrow d(\Xi_\beta\circ \mathcal G^{n+1}_Ep, \mathcal G^{n+1}_{E^*}\circ\Xi_\beta p)<\epsilon \quad\quad\qed\]}{}

The limitwise coarsenability of evolution theorem is very general. As we have not committed ourselves to any particular genomic data-structure the coarse-graining result we have obtained is applicable to any IPEA provided that it satisfies three abstract conditions: bounded thematic mean divergence, ambivalence, and non-departure. The fidelity of the coarse-graining depends on the the minimal bound on the thematic mean divergence. Maximum fidelity is achieved in the limit as this minimal bound tends to zero.

\section{Sufficient Conditions for Coarse-Graining IPGA Dynamics}

We now use the result in the previous section to argue that the dynamics of an IPGA with long genomes, uniform crossover, and fitness proportional selection can be coarse-grained with high fidelity for a relatively coarse schema partitioning, provided that the initial population satisfies a constraint called \emph{approximate schematic uniformity} and the fitness function satisfies a  constraint called \emph{low-variance schematic fitness distribution}. We stress at the outset that our argument is principled but informal, i.e. though the argument rests relatively straightforwardly on theorem 3, we do find it necessary in places to appeal to the reader's intuitive understanding of GA dynamics.

For any $n\in \mathbb Z^+$, let $\mathfrak B_n$ be the set of all bitstrings of length $n$. For some $\ell\gg1$ and some $m\ll \ell$, let $\beta:\mathfrak B_\ell\rightarrow \mathfrak B_m$ be some schema partitioning. Let $f^*:\mathfrak B_m\rightarrow \mathbb R^+$ be some function. For each $k\in \mathfrak B_m$, let  $D_k\in\Lambda^{\mathbb R^+}$ be some distribution over the reals with low variance such that the mean of distribution $D_k$ is $f^*(k)$. Let $f:\mathfrak B_\ell\rightarrow \mathbb R^+$ be a fitness function such that for any $k\in \mathfrak B_m$, the fitness values of the elements of $\langle k \rangle_\beta$ are independently drawn from the distribution $D_k$. For such a fitness function we say that fitness is \emph{schematically distributed with low-variance}.

Let $U$ be a set of distributions such that for any $k\in \mathfrak B_m$ and any $p\in U$, $\mathcal C_\beta(p,k)$ is approximately uniform. It is easily checked that $U$ satisfies the condition $\Xi_\beta(U)=\Lambda^{\mathfrak B_m}$. We say that the distributions in $U$ are \emph{approximately schematically uniform}.

Let $\delta$ be the minimal bound such that for all $p\in U$ and for all $k\in\mathfrak B_m$, $ |\mathcal E_f\circ\mathcal C_\beta(p,k)-f^*(k)|\leq\delta$. Then,  for any $\epsilon>0$, $\mathbf P(\delta< \epsilon) \rightarrow 1$ as $\ell-m\rightarrow\infty$. Because we have chosen $\ell$ and $m$ such that $\ell-m$ is `large', it is reasonable to assume that the minimal bound on the schematic mean divergence of $f$ on $U$ under $\beta$ is likely to be `low'.

Let $T\in \Lambda^{\mathfrak B_\ell}$ be a transmission function that models the application of uniform crossover. In sections 6 and 7 of \cite{interPopConstaints} we rigorously prove that a transmission function that models any mask based crossover operation is ambivalent under any schema partitioning. Uniform crossover is mask based, and $\beta$ is a schema partitioning, therefore $T$ is ambivalent under $\beta$.

Let $p_{\frac{1}{2}}\in \Lambda^{\mathfrak B_1}$ be such that $p_{\frac{1}{2}}(0)=\frac{1}{2}$ and $p_{\frac{1}{2}}(1)=\frac{1}{2}$. For any $p \in U$, $\mathcal S_fp$ may be `outside' $U$ because there may be one or more $k\in \mathfrak B_m$ such that $\mathcal C_\beta(\mathcal S_fp,k)$ is not quite uniform. Recall that for any $k\in \mathfrak B_m$ the variance of $D_k$ is low. Therefore even though $\mathcal S_fp$ may be `outside' $U$, the deviation from schematic uniformity is not likely to be large. Furthermore, given the low variance of $D_k$, the marginal distributions of $\mathcal C_\beta(\mathcal S_fp,k)$ will be very close to $p_{\frac{1}{2}}$. Given these facts and our choice of transmission function, for all $k\in K$, $\mathcal C_\beta(\mathcal V_T\circ\mathcal S_fp,k)$ will be more uniform than $\mathcal C_\beta(\mathcal S_fp,k)$, and we can assume that $\mathcal V_T\circ\mathcal S_fp$ is in $U$. In other words, we can assume that $E$ is non-departing over $U$.

Let $E=(\mathfrak B_\ell, T, f)$ and $E^*=(\mathfrak B_m, T^{\overrightarrow\beta}, f^*)$ be evolution machines. By the discussion above and the limitwise coarsenablity of evolution theorem one can expect that for any approximately thematically uniform distribution $p\in U$ (including of course the uniform distribution over $\mathfrak B_\ell$), the dynamics of $E^*$ when initialized with $\Xi_\beta p$ will approximate the projected dynamics of $E$ when initialized with $p$. As the bound $\delta$ is `low', the fidelity of the approximation will be `high'.

Note that the constraint that fitness be low-variance schematically distributed, which is required for this coarse-graining, is much weaker than the very strong constraint of schematic fitness invariance (all genomes in each schema must have the \emph{same} value) which is required to coarse-grain IPGA dynamics in \cite{conf/gecco/WrightVR03}.

\section{Conclusion}
It is commonly assumed that the ability to track the frequencies of schemata in an evolving infinite population across multiple generations under different fitness functions will lead to better theories of adaptation for the simple GA. Unfortunately tracking the frequencies of schemata in the naive way described in the introduction is computationally intractable for IPGAs with long genomes. A previous coarse-graining result \cite{conf/gecco/WrightVR03} suggests that tracking the frequencies of a family of low order schemata is computationally feasible, regardless of the length of the genomes, if fitness is schematically invariant (with respect to the family of schemata). Unfortunately this strong constraint on the fitness function renders this result useless if one's goal is to understand how GAs perform adaptation on real-world fitness functions.

In this paper we developed a simple yet powerful abstract framework for modeling evolutionary dynamics. We used this framework to show that the dynamics of an IPEA can be coarse-grained if it satisfies three abstract conditions. We then used this result to argue that the evolutionary dynamics of an IPGA with fitness proportional selection and uniform crossover can be coarse-grained (with high fidelity) under a relatively coarse schema partitioning if the initial distribution satisfies a constraint called approximate schematic uniformity (a very reasonable condition), and  fitness is low-variance schematically distributed. The latter condition is much weaker than the schematic invariance constraint previously required to coarse-grain selecto-mutato-recombinative evolutionary dynamics.

\vspace{.2cm}
\noindent \emph{Acknowledgements:} The reviewers of this paper gave me many useful comments, suggestions, and references. I thank them for their feedback. I also thank Jordan Pollack for supporting this work.
\bibliographystyle{plain}

\appendix \large \vspace{0.3cm}\noindent \textbf{Appendix} \small

\ifthenelse{\boolean{showProofs}}{

\begin{lemma}
For any finite set $X$, and any metric space $(\Upsilon,d)$,let  $\mathcal A:\Upsilon\rightarrow \Lambda^X$ and let $\mathcal B:X\rightarrow [\Upsilon\rightarrow [0,1]]$ be functions\footnote{For any sets $X, Y$ we use the notation $[X\rightarrow Y]$ to denote the set of all functions from $X$ to $Y$} such that for any $h\in \Upsilon$, and any $x\in X$, $(\mathcal B(x))(h)=(\mathcal A(h))(x)$. For any $x\in X$, and for any $h^*\in \Upsilon$, if the following statement is true
\begin{align*} \forall x\in X, \forall \epsilon_x>0, \exists \delta_x>0, \forall h\in \Upsilon, d(h,h^*)<\delta_x\Rightarrow |(\mathcal B(x))(h)-(\mathcal B(x))(h^*)|<\epsilon_x\end{align*}Then we have that \begin{align*}\forall \epsilon>0, \exists
\delta>0,  \forall h\in \Upsilon, d(h, h^*)<\delta \Rightarrow d(\mathcal A(h),\mathcal A(h^*))<\epsilon\end{align*}
\end{lemma}
This lemma says that $\mathcal A$ is continuous at $h^*$ if for all $x\in X$, $\mathcal B(x)$ is continuous at $h^*$.\\
\textsc{Proof:  } We first prove the following two claims

\begin{clam}
\begin{multline*}\forall x\in X \textrm{ s.t. } (\mathcal B(x))(h^*)>0,  \forall \epsilon_x>0, \exists \delta_x>0, \forall h\in \Upsilon,\\
    d(h,h^*)<\delta_x\Rightarrow |(\mathcal B(x))(h)-(\mathcal B(x))(h^*)|<\epsilon_x.(\mathcal B(x))(h^*)\end{multline*}
\end{clam}
This claim follows from the continuity of $\mathcal B(x)$ at $h^*$ for all $x\in X$ and the fact that $(\mathcal B(x))(h^*)$ is a positive constant w.r.t. $h$.
\begin{clam} For all $h\in  \Upsilon$
\[\sum_{\substack{x\in X \text{s.t.}\\(\mathcal A(h^*))(x)>\\(\mathcal A(h))(x)}}|(\mathcal A(h^*))(x)-(\mathcal A(h))(x)|=
 \sum_{\substack{x\in X \text{s.t.}\\(\mathcal A(h))(x)>\\(\mathcal A(h^*))(x)}}|(\mathcal A(h))(x)-(\mathcal A(h^*))(x)|\]
 \end{clam}
\noindent The proof of this claim is as follows: for all $h\in \Upsilon$,
{\allowdisplaybreaks\begin{align*}
&\sum_{x\in X}(\mathcal A(h^*)(x))-(\mathcal A(h))(x)=0\\
&\Rightarrow \sum_{\substack{x\in X \text{s.t.}\\(\mathcal A(h^*))(x)>\\(\mathcal A(h))(x)}}(\mathcal A(h^*))(x)-(\mathcal A(h))(x) -
 \sum_{\substack{x\in X \text{s.t.}\\(\mathcal A(h))(x)>\\(\mathcal A(h^*))(x)}}(\mathcal A(h))(x)-(\mathcal A(h^*))(x)=0\\
&\Rightarrow \sum_{\substack{x\in X \text{s.t.}\\(\mathcal A(h^*))(x)>\\(\mathcal A(h))(x)}}(\mathcal A(h^*))(x)-(\mathcal A(h))(x)=
 \sum_{\substack{x\in X \text{s.t.}\\(\mathcal A(h))(x)>\\(\mathcal A(h^*))(x)}}(\mathcal A(h))(x)-(\mathcal A(h^*))(x)\\
 &\Rightarrow \bigg|\sum_{\substack{x\in X \text{s.t.}\\(\mathcal A(h^*))(x)>\\(\mathcal A(h))(x)}}(\mathcal A(h^*))(x)-(\mathcal A(h))(x)\Bigg|=\Bigg|
 \sum_{\substack{x\in X \text{s.t.}\\(\mathcal A(h))(x)>\\(\mathcal A(h^*))(x)}}(\mathcal A(h))(x)-(\mathcal A(h^*))(x)\Bigg|\\
 &\Rightarrow \sum_{\substack{x\in X \text{s.t.}\\(\mathcal A(h^*))(x)>\\(\mathcal A(h))(x)}}|(\mathcal A(h^*))(x)-(\mathcal A(h))(x)|=
 \sum_{\substack{x\in X \text{s.t.}\\(\mathcal A(h))(x)>\\(\mathcal A(h^*))(x)}}|(\mathcal A(h))(x)-(\mathcal A(h^*))(x)|
\end{align*}
\noindent We now prove the lemma. Using claim 1 and the fact that $X$ is finite, we get that $\forall \epsilon>0$, $\exists \delta>0$, $\forall h\in[X\rightarrow\mathbb R]$ such that $d(h,h^*)<\delta$,
\begin{align*}
&\sum_{\substack{x\in X \text{s.t.}\\(\mathcal A(h^*))(x)>\\(\mathcal A(h))(x)}}|(\mathcal B(x))(h^*)-(\mathcal B(x))(h)|<\sum_{\substack{x\in X \text{s.t.}\\(\mathcal A(h^*))(x)>\\(\mathcal A(h))(x)}}\frac{\epsilon}{2}.(\mathcal B(x))(h^*)\\
&\Rightarrow \sum_{\substack{x\in X \text{s.t.}\\(\mathcal A(h^*))(x)>\\(\mathcal A(h))(x)}}|(\mathcal A(h^*))(x)-(\mathcal A(h))(x)|<\sum_{\substack{x\in X \text{s.t.}\\(\mathcal A(h^*))(x)>\\(\mathcal A(h))(x)}}\frac{\epsilon}{2}.(\mathcal A(h^*))(x)\\
&\Rightarrow \sum_{\substack{x\in X \text{s.t.}\\(\mathcal A(h^*))(x)>\\(\mathcal A(h))(x)}}|(\mathcal A(h^*))(x)-(\mathcal A(h))(x)|<\frac{\epsilon}{2}\quad\quad \qed
\end{align*}
\noindent By Claim 2 and the result above, we have that $\forall \epsilon>0$, $\exists \delta>0$, $\forall h\in[X\rightarrow\mathbb R]$ such that $d(h,h^*)<\delta$,
\begin{align*}
  \sum_{\substack{x\in X \text{s.t.}\\(\mathcal A(h))(x)>\\(\mathcal A(h^*))(x)}}|(\mathcal A(h))(x)-(\mathcal A(h^*))(x)|<\frac{\epsilon}{2}
\end{align*}
\noindent Therefore, given the two previous results, we have that $\forall \epsilon>0$, $\exists \delta>0$, $\forall h\in[X\rightarrow\mathbb R]$ such that $d(h,h^*)<\delta$,
\begin{align*}
&\sum_{x\in X}|(\mathcal A(h))(x)-(\mathcal A(h^*)(x))|<\epsilon \quad\quad \qed
\end{align*}}
\begin{lemma} \label{lemmaB}
Let $X$ be a finite set, and let $T\in \Lambda^X_m$ be a transmission function. Then for any $p'\in \Lambda^X$ and any $\epsilon>0$, there exists a $\delta>0$ such that for any $p\in \Lambda^X$,
\[d(p\,,\,p')<\delta\Rightarrow d(\mathcal V_Tp\,,\, \mathcal V_Tp')<\epsilon\]
\end{lemma}
\noindent \textit{Sketch of Proof:  } Let $\mathcal A:\Lambda^X\rightarrow\Lambda^X$ be defined such that $(A(p))(x)=(\mathcal V_Tp)(x)$. Let $\mathcal B:X\rightarrow[\Lambda^X\rightarrow [0,1]]$ be defined such that $(\mathcal B(x))(p)=(\mathcal V_Tp)(x)$. The reader can check that for any $x\in X$, $\mathcal B(x)$ is a continuous function. The application of lemma 1 completes the proof. \\

\noindent By similar arguments, we obtain the following two lemmas.
\begin{lemma} \label{lemmaC}
  Let $X$ be a finite set, and let $f:X\rightarrow \mathbb R^+$ be a function. Then for any $p'\in \Lambda^X$ and any $\epsilon>0$, there exists a $\delta>0$ such that for any $p\in \Lambda^X$,
  \[d(p\,,\,p')<\delta\Rightarrow d(\mathcal S_fp\,,\, \mathcal S_fp')<\epsilon\]
\end{lemma}

\begin{lemma} \label{lemmaD}
  Let $X$ be a finite set, and let $p\in \Lambda^X$ be a distribution. Then for any $f'\in [X\rightarrow \mathbb R^+]$, and any $\epsilon>0$, there exists a $\delta>0$ such that for any $f\in[X\rightarrow \mathbb R^+]$,
  \[d(f\,,\,f')<\delta\Rightarrow d(\mathcal S_fp\,,\, \mathcal S_{f'}p)<\epsilon\]
\end{lemma}}{}
\end{document}